**DL-EWF: Deep Learning Empowering Women's Fashion with Grounded-Segment-Anything Segmentation for Body Shape Classification**

Fatemeh Asghari, Mohammad Reza Soheili, Faezeh Gholamrezaie
Department of Computer Engineering, Faculty of Engineering, Kharazmi University, Tehran, Iran
Email : f.asghari@khu.ac.ir
Department of Electrical and Computer Engineering, Faculty of Engineering, Kharazmi University, Tehran, Iran
Email : soheili@khu.ac.ir
Department of Computer Science, Shahed University, Tehran, Iran
Email : faeze.gholamrezaie@shahed.ac.ir

**Abstract**
The global fashion industry plays a pivotal role in the global economy, and addressing fundamental issues within the industry is crucial for developing innovative solutions. One of the most pressing challenges in the fashion industry is the mismatch between body shapes and the garments of individuals they purchase. This issue is particularly prevalent among individuals with non-ideal body shapes, exacerbating the challenges faced. Considering inter-individual variability in body shapes is essential for designing and producing garments that are widely accepted by consumers. Traditional methods for determining human body shape are limited due to their low accuracy, high costs, and time-consuming nature. New approaches, utilizing digital imaging and deep neural networks (DNN), have been introduced to identify human body shape. In this study, the Style4BodyShape dataset is used for classifying body shapes into five categories: Rectangle, Triangle, Inverted Triangle, Hourglass, and Apple. In this paper, the body shape segmentation of a person is extracted from the image, disregarding the surroundings and background. Then, Various pre-trained models, such as ResNet18, ResNet34, ResNet50, VGG16, VGG19, and Inception v3, are used to classify the segmentation results. Among these pre-trained models, the Inception V3 model demonstrates superior performance regarding f1-score evaluation metric and accuracy compared to the other models.
**Keywords:** body shape classification, image segmentation, pre-trained models, human body shape, clothing style

## 1.Introduction

Clothing is a basic human necessity, and the world of online clothing stores is constantly expanding, with a seemingly endless selection of plus-size and petite clothing. However, one of the biggest challenges in this field is returning purchased clothes that don't fit. [1].
Finding clothes that fit well and make you feel good is essential for everyone, regardless of age, culture, or body shape [2]. However, this can be challenging, especially for people with unique body types. One way to address this challenge is to understand the different body shapes and how they relate to clothing fit. This information can be used to help people find clothes that flatter their natural figure and make them feel confident and comfortable. [3].

Research on body types in garment construction aims to identify the body parts that cause poor clothing fit. This information can then be used to modify patterns to create more flattering and comfortable garments [4].

In general, body measurements can be obtained using three methods: manual measurements, photography, and scanners.

Traditional methods of body measurement use manual 2D measurements and photographic images. However, manual measurement is prone to human error and difficulty in accurately determining the appropriate measurement points. 3D scanning technology provides a more accurate and repeatable method for obtaining body measurements, including dimensions that are difficult to measure manually, such as depth and angle [5].

3D body scanners are optical measurement systems that create digital copies of the human body's surface. They consist of light sources, cameras, software, and computers. The two main types of scanners are laser and structured light. In both systems, light is projected onto the person being scanned, and cameras detect the reflected light, which is distorted by the person's body shape. The displacement of light is used to calculate the coordinates of 3D points on the body [6].

After obtaining the required measurements, four methods exist for classifying body shapes into different classes. These methods include the drop value, principal component analysis, the ratio of body sizes, and neural networks. The drop value refers to the difference between the size of the chest circumference and the hip circumference, and the chest circumference and the waist circumference.

The second method, principal component analysis (PCA) with clustering, reduces the dimensionality of the body measurement data and then uses clustering algorithms to classify body shapes into different categories. The third method combines the first two methods. The last method uses neural networks to classify body shapes, which has higher accuracy than other methods.

The classification results show that the most common body shapes fall into five categories: Rectangle, Triangle, Inverted Triangle, Hourglass, and Apple. Human body shape is a complex concept, and accurately classifying it requires complex methods. Neural networks have outperformed other methods, such as statistical and traditional methods, due to their ability to process complex data and learn complex models. Neural networks can also process data in parallel, making them well-suited for large-scale problems. In the fashion industry, where consumer demand constantly changes, neural networks are a promising tool for classifying body shapes. However, training neural networks for this task requires high-quality labeled data.

In this paper, we propose a body shape classification model for two-dimensional images robust to environmental changes, lighting changes, the presence of other objects in the image, background changes, and different viewpoints. The model takes an image of a person as input, regardless of the environment they are in, and outputs a mask of the person's body. This mask is then given as input to a classifier to identify the person's body shape.

Our model is unique in several ways. First, it is robust to environmental changes. This means it can accurately identify body shapes even in challenging conditions such as complex environments, varying lighting, and the presence of other objects in the image. This is in contrast to previous approaches, which were often sensitive to environmental changes and could produce inaccurate results. Second, our model uses a segmentation model to extract the person's mask. This allows the model to focus on the person and ignore the background. This is more accurate and efficient than previous approaches to whole-image classification. Next, the segmentation outcomes are classified using different pre-trained models, including ResNet18, ResNet34, ResNet50, VGG16, VGG19, and Inception v3. Finally, this model's performance is evaluated using metrics such as f1-score and accuracy. Based on the research findings, the best pre-trained model for classifying body shapes is the Inception V3 model.

The rest of the paper is organized as follows: Section 2 presents related work in this paper. Section 3 describes the body type, data set, and proposed model. Section 4 introduces the experimental results. Section 5 introduces the comparison methods. Finally, section 6 concludes the paper.

## 2.Related Work

To classify the body shapes, the body measurements must be obtained first, and then the body shape is determined using these measurements. There are three main methods for obtaining body measurements: manual measurement, photography, and 3D scanning. Manual measurement is the most traditional method. A trained professional uses a tape measure to measure the person's body. This method is accurate, but it can be time-consuming and labor-intensive. Photography is a more efficient way to obtain body measurements [7]. A person is photographed from multiple angles, and then a computer vision algorithm extracts the measurements from the images. This method is less accurate than manual measurement, but it is faster and easier to use [8]. 3D scanning is the most accurate method for obtaining body measurements. A 3D scanner scans the person's body, and then a computer algorithm extracts the measurements from the scan data. This method is the most expensive of the three, but it is also the most accurate [9].

Classifying body shape is a challenging task with many applications in the fashion industry. There are a variety of methods that can be used, each with its advantages and disadvantages. The four main methods for classifying body shape are drop value, PCA with clustering, the combination of the first two methods, and neural network.

### 2.1 Drop Value

In the first method, body shapes can be classified using the drop value. The drop value is the difference between the chest circumference and the hip circumference, and the difference between the chest circumference and the waist circumference. Makhanya et al. collected measurements from 234 African and Caucasian women aged 18 to 25 using a 3D scanner. The scanner extracted perimeter, width, and height measurements from the chest, abdomen, waist, hips, pelvis, and thighs. In this study, the drop value was used to identify and distinguish between body shapes. For each drop value, a range was defined that included the minimum and maximum values, the mean, and the standard deviation. The difference in size between the hips and chest was used to determine the Triangular and inverted Triangular shapes, and the difference in size between the chest and waist was used to identify the Hourglass, Apple, and Rectangular shapes. Specifically, if the difference in size between the hips and chest is in the range (Mean, Max), the body shape is Triangular. If the difference in size between these two is less than zero, meaning that the chest size is larger than the hip size, the body shape is inverted Triangular. If the difference in size between the chest and waist is in the range (Mean, Max), the body shape is the Hourglass. If the difference in size between these two is in the range (Mean, Mean - 3 Standard Deviation), the body shape is Rectangular. And if the difference in size between these two is in the range (Min, -3 Standard Deviation), the body shape is Apple [10].

In the paper by Parker et al., the effects of body measurement methods on body shape classification using the FFIT system (which uses the drop value method) are investigated. The SizeUK survey, the first national survey of the British population in the 1950s, collected 3D body measurements of the population using the same scanner used in FFIT, to update sizing charts. A comparison of SizeUK with FFIT shows that FFIT uses different definitions for obtaining the size of a region than SizeUK. For example, to obtain the hip circumference, FFIT considers the largest circumference from the waist to the groin as the hip circumference, while SizeUK considers the prominence of the hip or buttocks as the hip circumference. This difference in the definitions of body measurement sections leads to misclassification of body shape. Using Cohen's kappa test, the agreement between the classification of body shapes under different measurement definitions is measured. The results show that the upper Hourglass, inverted Triangle, and Rectangle groups are still correctly classified even with different measurements, while the Hourglass, Triangle, lower Hourglass, and spoon groups are misclassified with different definitions [11].

### 2.2 PCA and Clustering

In the second method, classification is performed using principal component analysis (PCA) and clustering. After obtaining body measurements, the dimensions of the data are first reduced using PCA. Then, clustering is performed on the reduced data using clustering algorithms such as k-means and k-means++.

In their study, Kim, Song, and Ashdown divided the bodies of petite women into different groups based on their body shape. The dataset consisted of 2714 three-dimensional measurements of American women aged 18 to 35. Women with a height of less than 4/5 inches were considered petite. As a result, only the measurements for these women were kept, which numbered 1618. Using PCA, the number of variables was reduced. The X-means algorithm a hierarchical method, was then used to determine the number of clusters by the algorithm. Then, the k-means algorithm was used to cluster the data with values of two to five as the number of clusters. By comparing the two methods, the final number of clusters was set to four, and petite women were divided into four clusters: women with small upper bodies, women with small lower bodies, petite women, and women with large sizes [12].

In the paper by Wang et al., three-dimensional data of 333 randomly selected Chinese women were collected using a scanner to be used for classifying their lower body. In general, 24 body variables that affect the shape of the lower body were used, and by applying the PCA algorithm to it, the number of components is reduced to three. These three factors include the width measurements of the lower body, such as waist circumference and hip width, the height of the individual, and the measurements related to the groin. Then, using the k-means algorithm and setting k to three, the data are clustered. The first cluster is people who are tall and have a fat body; people in the second cluster have a short height and an average body, and people in the third cluster have a tall height and a thin body [13].

### 2.3 Combination of drop value, PCA, and clustering

The third method for body shape classification combines the first two methods or uses the ratio between body measurements instead of the drop value. This is because the ratio between body measurements is more effective for classification than the original measurements. In this way, after obtaining the body measurements, some of the ratios such as the ratio of chest circumference to waist circumference, the ratio of upper body length to chest size, and others are calculated. In the next step, the dimensions of the data are reduced using the PCA algorithm, and then the clustering algorithm is used on the data whose dimensions have been reduced [14].

In the paper by Dong, Zeng, and Koehl, a dynamic fuzzy clustering method is proposed to identify the lower body shape of a specific population for the development of pants design. Since only a few body measurements, called key dimensions, are important for a specific population and a specific clothing style, for pants design, the key dimensions related to the lower body shape should be selected, which are: waist shape, hip shape, stomach shape, thigh shape, and calf shape. Because in clothing design, the differences or ratios between body measurements are generally more important than direct measurements for classifying body shapes, the ratio between different measurements is calculated for each individual. To cluster the data, a preprocessing is first performed on them to remove outliers, then the measurements for each individual are normalized, and in the final step, the data are divided into five different clusters using the fuzzy clustering algorithm [15].

Cui's study collected 15 measurements of the upper and lower body of 600 women aged 18 to 23, including height, chest circumference, waist circumference, hip circumference, shoulder width, and shoulder angle. To explore the relationships between these measurements, 13 variables were calculated that measure the ratio of some of the measurements to each other, such as the ratio of chest circumference to waist circumference, the ratio of upper body length to chest circumference, the ratio of hip circumference to waist circumference, and the ratio of lower body length to hip circumference. Finally, the upper and lower body of each individual were classified separately using these 13 variables. Factor analysis was used to measure the impact of each of the 13 variables on the classification of the upper and lower body of individuals. The results showed that the ratio of upper body length to horizontal chest length and the ratio of upper body length to waist circumference have the greatest impact on the classification of the upper body. Using the k-means algorithm, the upper body of individuals was placed into five clusters. Similarly, the difference in the size of the hip circumference with the waist circumference and the ratio of the size of these two have the greatest impact on the classification of the lower body. Using the k-means algorithm, the lower body of individuals was placed into four clusters [16].

**2.4 neural networks**

Neural networks are powerful computational models that can be used to classify body shapes with greater accuracy and effectiveness than traditional methods. They work by processing and analyzing body shape data using complex mathematical algorithms. This is especially useful in the fashion industry, where body shape classification is often complex and variable. Neural networks are particularly well-suited for handling uncertain and nonlinear data, which is common in the fashion industry. For example, neural networks can be used to classify body shapes based on a variety of features, such as height, weight, measurements, and even personal preferences. This level of flexibility and adaptability is not possible with traditional methods. Four papers have been published in the field of body shape classification using neural networks.

- The first paper uses discriminative analysis to extract features that distinguish body shapes. These features are then used as input to a three-layer perceptron neural network for classification [17].

- The second paper introduces a method that can accurately identify body types using label filtering and pseudo-feature synthesis modules. Initially, body shape features are extracted using anthropometric measurements. Then, noisy labels are identified and removed using a label filtering module to ensure the network is trained on correct labels. Finally, by utilizing pseudo-feature synthesis, the data is balanced in the feature space among classes, and a classifier is trained to recognize body types. The results of experiments conducted on a dataset collected from online feeds demonstrate that the proposed method outperforms existing methods in terms of accuracy [18].
- The third paper constructs a dataset of 80 background-removed images representing four body shapes: Hourglass, Triangle, inverted Triangle, and Rectangle. The images are resized to a uniform size of 300x300 pixels and edge features are extracted using the Sobel filter algorithm. A Gaussian filter is then applied to the extracted edges to reduce noise. The preprocessed images are then used as input to an AlexNet convolutional neural network (CNN), which is trained on the preprocessed images for nine iterations and then tested on a set of test images [19].
- The fourth paper proposes a method for classifying body shape using a single image and the user's height. The approach involves a pipeline of deep learning models trained on benchmarking datasets for instance segmentation and keypoint estimation. The pipeline includes background subtraction, keypoint estimation, and distance calculations. The estimated locations and distances are then used to classify the body shape into one of five possible classes [20].

## 3.Method

### 3.1 Body Shape Types

Choosing clothes that flatter your body shape is an essential part of personal style. It can make you look and feel your best. Understanding your body shape can make shopping for clothes easier and more enjoyable. It can also help you find clothes that make you look and feel confident. Figure 1 shows various body shapes, categorized into five primary groups.

- **Hourglass**: People with an Hourglass body shape have a well-proportioned figure with a defined waist and similar-sized bust and hips.
- **Apple**: Apple-shaped bodies have a larger waist and smaller bust and hips. People with this body shape may have more body fat in the abdominal region.
- **Rectangle**: People with a Rectangle body shape have a straight up and down figure, with minimal difference between their bust, waist, and hip measurements.
- **Inverted Triangle**: this body shape is characterized by a bust that is larger than the hips. This means that people with this body shape have broad shoulders and a bust, and narrow hips.
- **Triangle:** The Triangle body shape is characterized by hips that are wider than the bust.

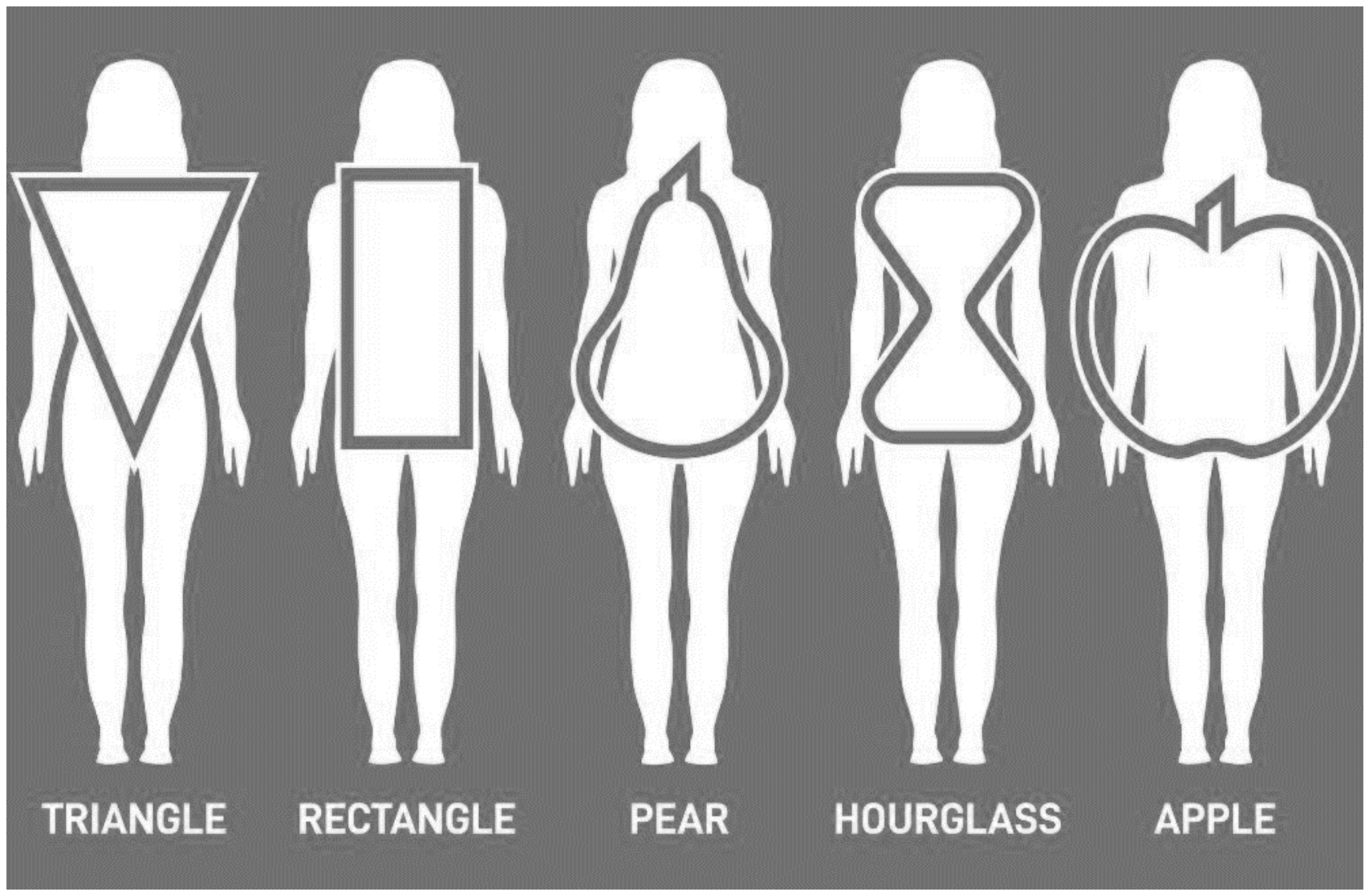

Figure 1. Body shape types [21]

### 3.2 Body Shape Dataset

The dataset used in this study is a subset of the Style4BodyShape dataset, which contains 349,000 images of 270 women in various outfits. The outfits are categorized into five main classes: dresses, pants, skirts, tops, and outerwear [22]. To clean the dataset, only images in which the subject is wearing form-fitting clothing and their hands are separated from their bodies are retained. This ensures that the key measurements used to define body shape, such as bust circumference, waist circumference, and hip circumference, can be clearly identified. After cleaning and labeling images according to the definitions provided in the previous section, the number of images in different categories, namely Apple, Hourglass, Inverted Triangle, Rectangle, and Triangle, becomes 50, 315, 166, 315, and 95, respectively. To increase the number of images in each class to 1000, data augmentation techniques such as rotation and flipping are used.

### 3.3 AI Body Measurement

The DL-EWF model is composed of two parts. In the first part, a body segmentation model called Grounded-Segment-Anything is used to extract the body segmentation mask from the image. In the next step, the body segmentation mask is given as input to a classification network to determine the body shape. The types of classifiers used in this study are ResNet18, ResNet34, ResNet50, VGG16, VGG19, and Inception. Figure 2 shows the view of the DL-EWF model.

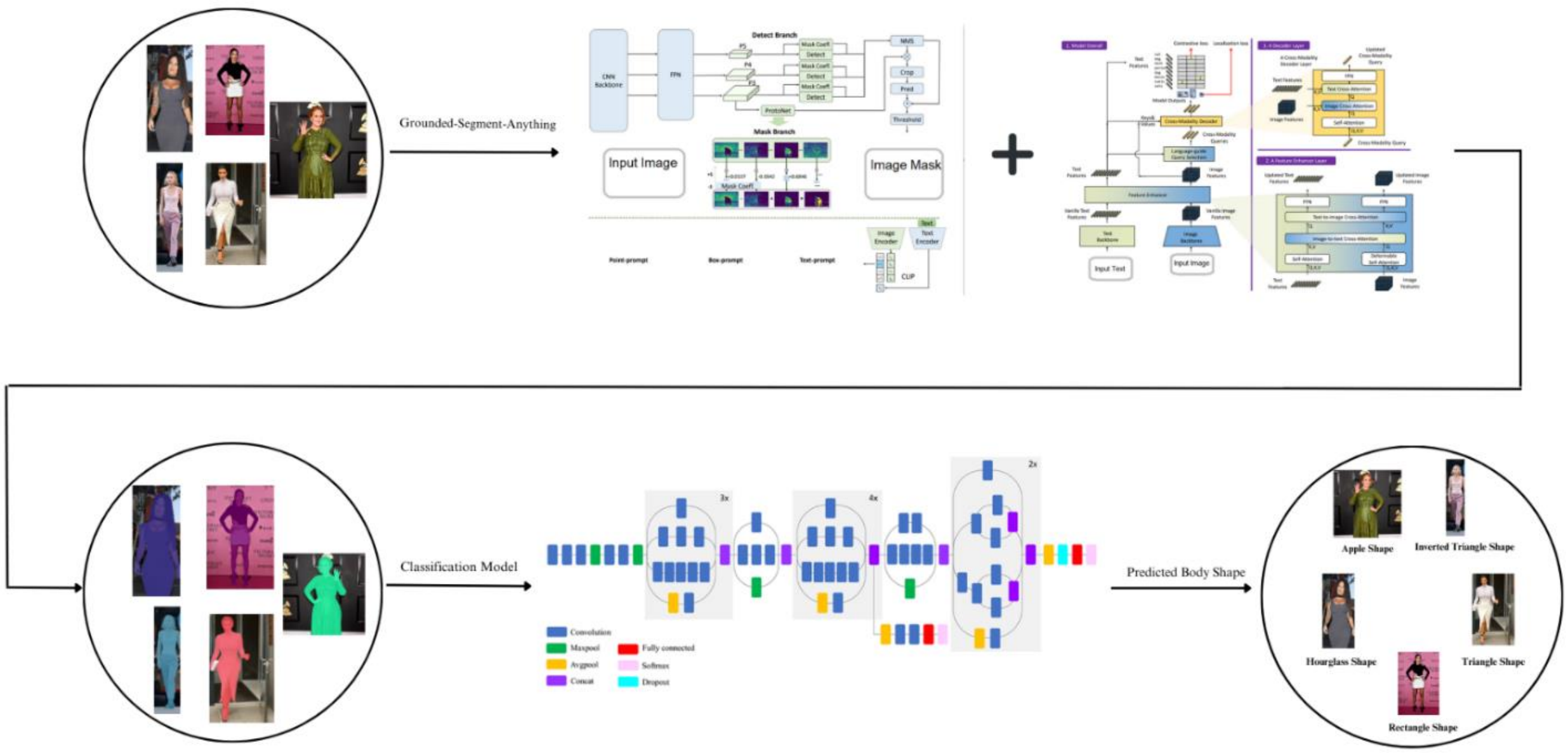

Figure 2. DL-EWF model architecture

To find the person in an image for body shape classification, we use a segmentation model called Grounded-Segment-Anything. This model is a combination of two other models: Grounding DINO and Segment Anything. Grounded-Segment-Anything can identify and segment objects in images based on text, bounding box, or point information [23].

In Grounded-Segment-Anything, a combination of two networks, Grounding DINO and Segment Anything, is used. The Grounding DINO model is an advanced object detection model that improves its performance by using image and text information. It uses a dual-encoder single-decoder architecture. One encoder is used to extract features from images and the other encoder is used for text. Then, the decoder is used for prediction [24]. The Fast Segment Anything model is used for image segmentation. This model uses an object detection based model such as YOLOv8-seg. In this model, all segmentations in the image are first extracted. After segmenting all objects in an image using YOLOv8, the second stage is to use point, box, or text prompts to identify specific objects of interest by matching selected points, bounding boxes, or text embeddings to masks, merging multiple masks, or selecting the mask with the highest similarity score [25].

Grounded-Segment-Anything can segment any object in an image, given a text prompt. To extract body shape segmentation, we simply give the model an image of a person and the text prompt "Person". Figure 3 shows the output results of Grounded-Segment-Anything for the five types of body shapes in the Style4BodyShape dataset.

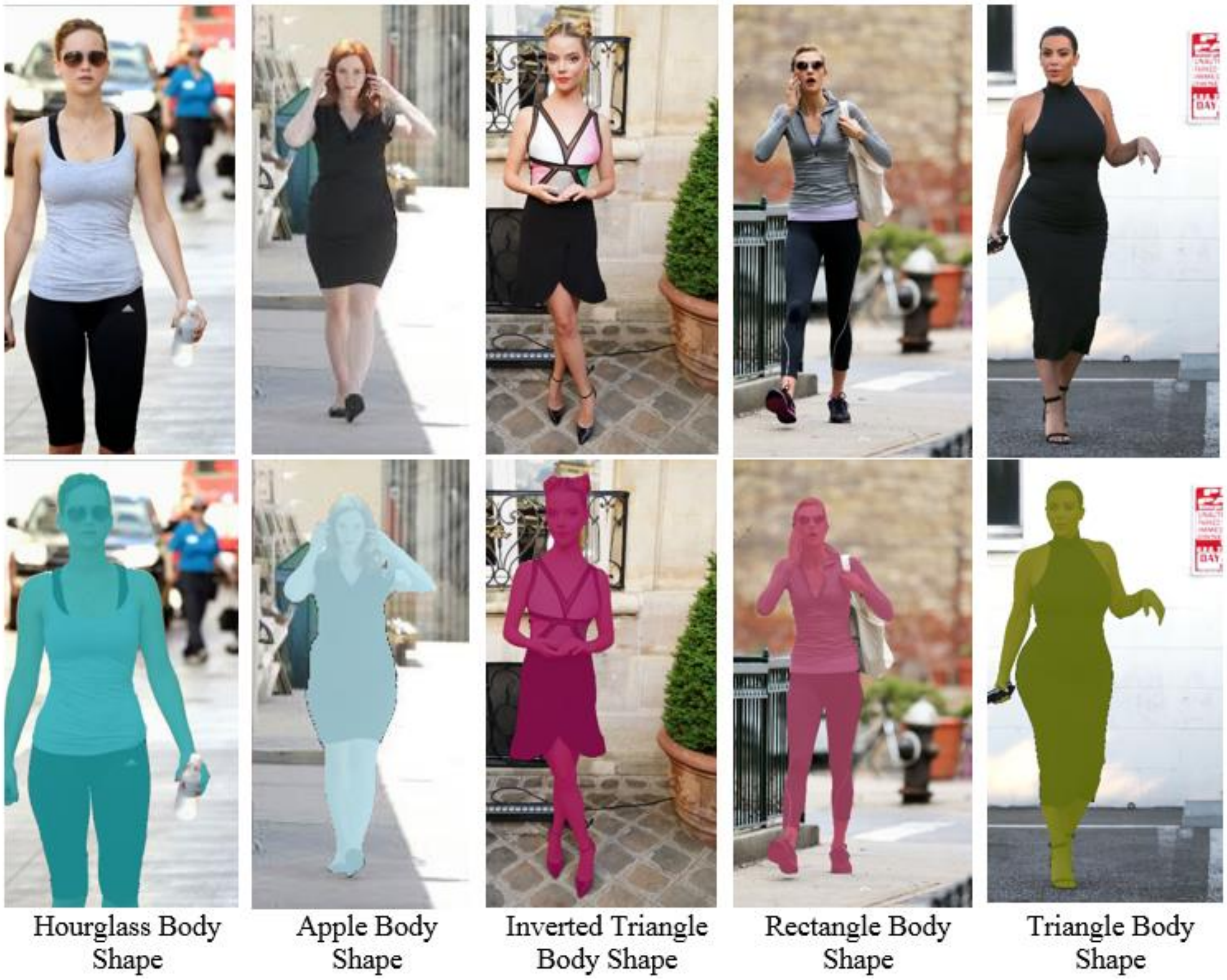

Figure 3. Body shape segmentation results for five types of body shapes in the Style4BodyShape dataset

## 4.EXPERIMENTAL RESULTS

Deep Learning has revolutionized image processing with the development of powerful convolutional neural networks (CNNs). CNNs can perform a variety of image-processing tasks, including image classification, with high accuracy. ResNet, VggNet, and Inception v3 are some of the most popular CNN architectures that have been shown to achieve state-of-the-art results on a variety of image classification benchmarks.

Segmented images, generated by partitioning an image into pixel-level segments with assigned labels, facilitate diverse tasks like object detection, scene understanding, and image analysis. Classifying these segments presents a significant challenge, requiring models to learn the intricacies of visual features associated with each object or region in the image.

ResNet18, ResNet34, ResNet50, VGG16, VGG19, and Inception v3 are all CNN architectures that have been shown to be effective for classifying segmented images. These architectures have been trained on large datasets of segmented images, and they have learned to extract the visual features that are relevant for classifying different objects or regions.

Here are some of the advantages of using these CNNs for classifying segmented images:

- These architectures are well-established and have been shown to achieve state-of-the-art results on a variety of image classification benchmarks.
- These architectures are relatively easy to train and deploy.
- There are pre-trained models available for these architectures, which can be used as a starting point for training a model on a specific dataset of segmented images.

Fine-tuning is a process where a pretrained model is adapted to a specific task by generalizing from task-oriented data. This process is performed by adjusting the weights of the model using a new task-specific dataset. As a result, the model, focusing on specific task data, has better capability in producing accurate and reasonable outputs. The process of fine-tuning models and outputs produced by the aforementioned models are presented below:

- **VGG16 :** All layers except the last five of VGG1 were frozen, and the model was trained, but it exhibited underfitting.
- **VGG19 :** Despite fine-tuning only the last twelve layers of the VGG19 pre-trained network, it still resulted in underfitting when classifying body shapes in segmented images.
- **ResNet18 :** Focusing modifications on the fourth stage layers of the ResNet18 [26] pre-trained architecture significantly improved body shape classification accuracy compared to the previous models.
- **ResNet34 :** The ResNet34 architecture draws inspiration from the Residual Network design and is composed of 34 layers, consisting of convolutional layers and residual blocks. In terms of classifying body shapes, this network outperforms previous networks.
- **ResNet50 :** This network, similar to the previous models, belongs to the ResNet family but comprises 50 layers. Unfortunately, it doesn't exhibit strong performance on the test data, with a maximum accuracy of 39% and an f1-score of 34%.
- **Inception v3 :** The Inception V3 [27] network was trained from scratch on the target dataset, without using any prior weights. It performs the best among all previous networks on the segmentation dataset, achieving the highest accuracy and f1-score.

The performance of the mentioned models in Table 1 has been compared using the confusion matrix, classification report, training vs validation loss.

**Table 1. Comparing the performance of different models in body shape classification**

| Model Name | Confusion Matrix | Classification Report | Loss vs Epoch |
|---|---|---|---|

## VGG16

Confusion Matrix (Actual Labels; Predicted Lables)

| | ○ | ⧗ | ▽ | ▯ | △ |
|---|---|---|---|---|---|
| ○ | 2 | 13 | 0 | 0 | 0 |
| ⧗ | 0 | 141 | 0 | 0 | 0 |
| ▽ | 0 | 0 | 0 | 0 | 0 |
| ▯ | 0 | 0 | 0 | 0 | 0 |
| △ | 0 | 0 | 0 | 0 | 0 |

Classification Report

| | precision | recall | f1-score | support |
|---|---|---|---|---|
| 0 | 0.67 | 0.12 | 0.20 | 17 |
| 1 | 0.41 | 1.00 | 0.58 | 141 |
| 2 | 0.00 | 0.00 | 0.00 | 59 |
| 3 | 0.00 | 0.00 | 0.00 | 112 |
| 4 | 0.00 | 0.00 | 0.00 | 19 |
| accuracy | | | 0.41 | 348 |
| macro avg | 0.22 | 0.22 | 0.16 | 348 |
| weighted avg | 0.20 | 0.41 | 0.24 | 348 |

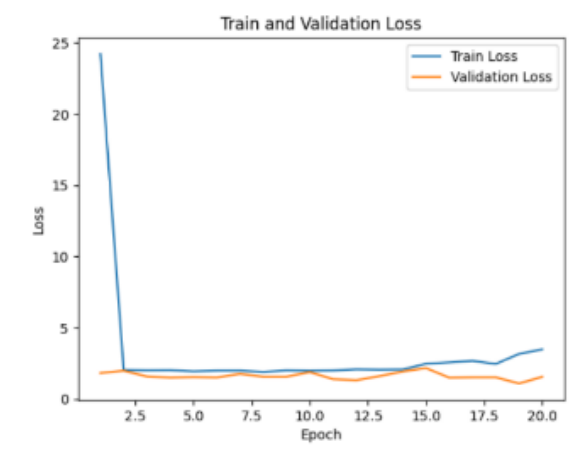

## VGG19

Confusion Matrix (Actual Labels; Predicted Lables)

| | ○ | ⧗ | ▽ | ▯ | △ |
|---|---|---|---|---|---|
| ○ | 0 | 0 | 0 | 0 | 0 |
| ⧗ | 11 | 11 | 12 | 4 | 2 |
| ▽ | 0 | 14 | 45 | 0 | 0 |
| ▯ | 0 | 0 | 0 | 0 | 0 |
| △ | 0 | 2 | 12 | 2 | 4 |

Classification Report

| | precision | recall | f1-score | support |
|---|---|---|---|---|
| 0 | 0.00 | 0.00 | 0.00 | 17 |
| 1 | 0.37 | 0.08 | 0.13 | 141 |
| 2 | 0.19 | 0.76 | 0.31 | 59 |
| 3 | 0.00 | 0.00 | 0.00 | 112 |
| 4 | 0.05 | 0.21 | 0.08 | 19 |
| accuracy | | | 0.17 | 348 |
| macro avg | 0.12 | 0.21 | 0.10 | 348 |
| weighted avg | 0.18 | 0.17 | 0.11 | 348 |

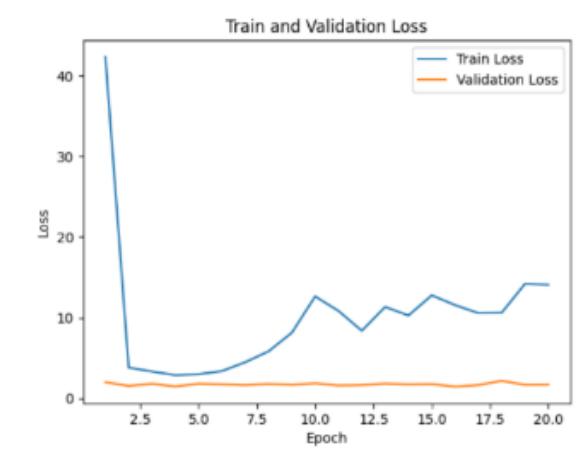

## ResNet18

Confusion Matrix (Actual Labels; Predicted Lables)

| | ○ | ⧗ | ▽ | ▯ | △ |
|---|---|---|---|---|---|
| ○ | 8 | 5 | 1 | 2 | 1 |
| ⧗ | 25 | 79 | 18 | 14 | 5 |
| ▽ | 5 | 32 | 6 | 14 | 2 |
| ▯ | 2 | 23 | 5 | 63 | 19 |
| △ | 0 | 2 | 1 | 9 | 5 |

Classification Report

| | precision | recall | f1-score | support |
|---|---|---|---|---|
| 0 | 0.57 | 0.47 | 0.52 | 17 |
| 1 | 0.50 | 0.56 | 0.53 | 141 |
| 2 | 0.24 | 0.10 | 0.14 | 59 |
| 3 | 0.56 | 0.56 | 0.56 | 112 |
| 4 | 0.16 | 0.32 | 0.21 | 19 |
| accuracy | | | 0.47 | 348 |
| macro avg | 0.41 | 0.40 | 0.39 | 348 |
| weighted avg | 0.46 | 0.47 | 0.45 | 348 |

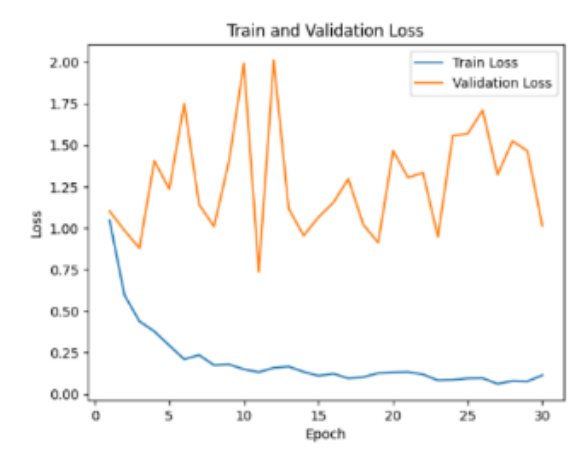

## ResNet34

Confusion Matrix (Actual Labels; Predicted Lables)

| | ○ | ⧗ | ▽ | ▯ | △ |
|---|---|---|---|---|---|
| ○ | 9 | 4 | 2 | 0 | 2 |
| ⧗ | 2 | 68 | 32 | 27 | 12 |
| ▽ | 0 | 19 | 40 | 16 | 5 |
| ▯ | 1 | 29 | 19 | 63 | 0 |
| △ | 0 | 8 | 6 | 2 | 3 |

Classification Report

| | precision | recall | f1-score | support |
|---|---|---|---|---|
| 0 | 1.00 | 0.53 | 0.69 | 17 |
| 1 | 0.52 | 0.48 | 0.50 | 141 |
| 2 | 0.48 | 0.19 | 0.27 | 59 |
| 3 | 0.46 | 0.68 | 0.55 | 112 |
| 4 | 0.23 | 0.26 | 0.24 | 19 |
| accuracy | | | 0.49 | 348 |
| macro avg | 0.54 | 0.43 | 0.45 | 348 |
| weighted avg | 0.50 | 0.49 | 0.47 | 348 |

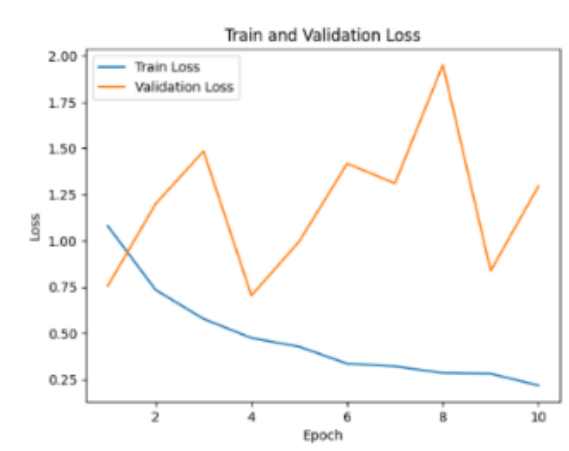

## ResNet50

Confusion Matrix (Actual Labels; Predicted Lables)

| | ○ | ⧗ | ▽ | ▯ | △ |
|---|---|---|---|---|---|
| ○ | 12 | 2 | 1 | 1 | 1 |
| ⧗ | 14 | 15 | 38 | 60 | 14 |
| ▽ | 15 | 13 | 22 | 8 | 1 |
| ▯ | 2 | 6 | 18 | 82 | 4 |
| △ | 3 | 4 | 4 | 2 | 4 |

Classification Report

| | precision | recall | f1-score | support |
|---|---|---|---|---|
| 0 | 0.27 | 0.71 | 0.39 | 17 |
| 1 | 0.57 | 0.11 | 0.19 | 141 |
| 2 | 0.30 | 0.37 | 0.33 | 59 |
| 3 | 0.43 | 0.73 | 0.54 | 112 |
| 4 | 0.36 | 0.21 | 0.27 | 19 |
| accuracy | | | 0.39 | 348 |
| macro avg | 0.39 | 0.43 | 0.34 | 348 |
| weighted avg | 0.45 | 0.39 | 0.34 | 348 |

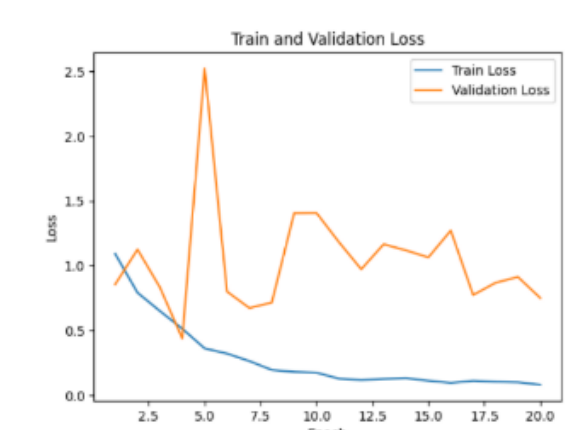

## Inception v3

Confusion Matrix (Actual Labels; Predicted Lables)

| | ○ | ⧗ | ▽ | ▯ | △ |
|---|---|---|---|---|---|
| ○ | 12 | 4 | 1 | 0 | 0 |
| ⧗ | 15 | 75 | 35 | 14 | 2 |
| ▽ | 4 | 31 | 13 | 9 | 2 |
| ▯ | 1 | 17 | 8 | 80 | 6 |
| △ | 0 | 3 | 4 | 7 | 7 |

Classification Report

| | precision | recall | f1-score | support |
|---|---|---|---|---|
| 0 | 0.57 | 0.71 | 0.63 | 17 |
| 1 | 0.62 | 0.53 | 0.57 | 141 |
| 2 | 0.30 | 0.22 | 0.25 | 59 |
| 3 | 0.56 | 0.71 | 0.62 | 112 |
| 4 | 0.33 | 0.37 | 0.35 | 19 |
| accuracy | | | 0.53 | 348 |
| macro avg | 0.48 | 0.51 | 0.49 | 348 |
| weighted avg | 0.53 | 0.53 | 0.53 | 348 |

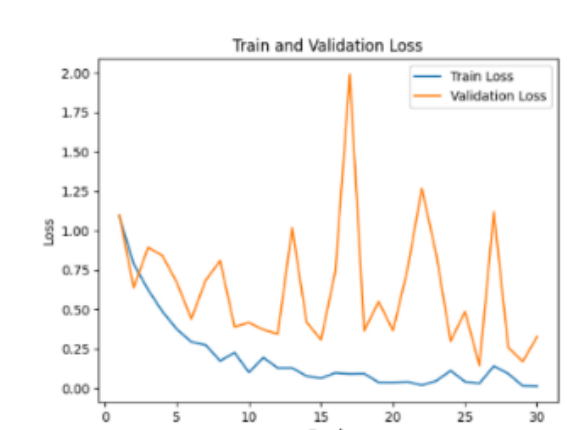

By analyzing the performance of 5 convolutional neural networks for body shape classification, their weaknesses, and strengths are identified. VGG16 only reliably identified Hourglass and Apple shapes, while VGG19 completely failed to discriminate between different types. Both models exhibited underfitting, as evidenced by their inability to differentiate shapes beyond basic categories and the error plots showing poor validation performance alongside low training loss.

ResNet18 model excelled at classifying Hourglass and Rectangular shapes (label 1 and 3). However, it misclassified inverted Triangle shapes (label 4) as Hourglass due to their narrow waists (confusion matrix). Its error plot indicated fluctuating validation errors despite decreasing training loss.

ResNet34 achieving the highest accuracy for circular (label 0) and Rectangular shapes (label 3), this model showed broader capability than VGG and ResNet18 (confusion matrix). However, it struggled with Triangular shapes (label 4). With a 49% accuracy and 45% F1-score on test data, further optimization is needed. Its error plot suggested faster F1-score improvement with fewer epochs.

ResNet50 While performing better than other models for labels 0, 2, and 3, it still confused Hourglass (label 1) and Rectangular shapes (label 3). Its limited test data performance peaked at 39% accuracy and 34% F1-score.

Inception V3 Demonstrating the strongest performance on test data, this model excelled at predicting labels 0, 1, and 3 but underperformed for Triangle and inverted Triangle shapes (labels 2 and 4). Notably, it achieved the highest accuracy and F1-score on the segmented dataset compared to all previous models (error plot and confusion matrix).

## 5. Comparing Methods

**Table 2.** Comparison of related work with proposed method

| Paper | Number of Classes | Measurements | Person Segmentation |
|---|---|---|---|
| Kim *et* al. [17] | 3 | ✓ | × |
| Body Shape Calculator [18] | 5 | ✓ | × |
| SmartFit [19] | 4 | × | × |
| Trotter *et* al. [20] | 5 | ✓ | ✓ |
| DL-EWF (Ours) | 5 | × | ✓ |

In contrast to previous works such as Kim et al., Body Shape Calculator, and Trotter et al., our proposed method does not rely on any body measurements. Similar to Trotter et al., our method utilizes instance segmentation to remove the background, allowing for the extraction of the person regardless of environmental changes, lighting variations, presence of other objects, background variations, and scenery. However, there is a difference between our method and Trotter et al. in terms of the training dataset. While Trotter et al. used data of individuals standing in a natural pose for training, our training dataset, Style4bodyshape, consists of images of famous individuals in various poses. This dataset compilation poses a challenge in accurately classifying body shapes due to the variations in body postures.

## 6. Conclusion

This paper proposes a novel body shape classification method that outperforms existing methods in several aspects. The method uses a segmentation model to extract a person mask, which focuses on the person of interest and ignores the background. It also has environmental robustness, which allows for body shape classification in challenging conditions.

The proposed method consists of two stages. In the first stage, body segmentation is performed using the Grounded-Segment-Anything model. This model extracts body segmentation regardless of environmental and perspective changes, lighting changes, and the presence of other objects in the image. In the second stage, the segmentation is used as input to a variety of models, including ResNet18, ResNet34, ResNet50, VGG16, VGG19, and Inception v3, for classification. Inception v3 achieved the highest accuracy and F1 score on the test set.

To improve the performance of neural network-based models for human body shape identification and classification, several approaches can be considered. One approach is to incorporate virtual reality (VR) and augmented reality (AR) capabilities into body shape recognition models. This would allow customers to try on women's clothing on their own images and see how the clothing would look on their bodies before purchasing. A second approach is to collect more data with labels provided by experts. This process would increase the diversity and distribution of the data and help models learn more accurate patterns for identifying and classifying body shapes. With the collection and labeling of data, pre-trained models for body shape classification can be created. These actions could improve aspects such as the online shopping experience and more accurate human body shape detection.